\documentclass[10pt,twocolumn]{ICCAS2024}
 
\usepackage{diagbox}
\usepackage{amssymb}
\usepackage{url}
\usepackage{amsmath,bm}
\usepackage{subcaption}

\usepackage{multirow}
\usepackage{tabularx}
\usepackage{longtable}
\usepackage{booktabs}
\usepackage{multicol}

\DeclareMathOperator*{\argmin}{argmin}

\begin{document}

\title{Enhancing Social Robot Navigation with Integrated Motion Prediction and Trajectory Planning in Dynamic Human Environments}

\author{ Thanh Nguyen Canh${}^{1,2}$, Xiem HoangVan${}^{2*}$ and Nak Young Chong${}^{1}$}

\affils{ ${}^{1}$School of Information Science, Japan Advanced Institute of Science and Technology \\
Ishikawa 923-1292, Japan (\{thanhnc, nakyoung\}@jaist.ac.jp) \\
${}^{2}$University of Engineering and Technology, Vietnam National University \\
Hanoi 10000, Vietnam (xiemhoang@vnu.edu.vn) {\small${}^{*}$ Corresponding author}\\
}


\abstract{
    Navigating safely in dynamic human environments is crucial for mobile service robots, and social navigation is a key aspect of this process. In this paper, we proposed an integrative approach that combines motion prediction and trajectory planning to enable safe and socially-aware robot navigation. The main idea of the proposed method is to leverage the advantages of Socially Acceptable trajectory prediction and Timed Elastic Band (TEB) by incorporating human interactive information including position, orientation, and motion into the objective function of the TEB algorithms. In addition, we designed social constraints to ensure the safety of robot navigation. The proposed system is evaluated through physical simulation using both quantitative and qualitative metrics, demonstrating its superior performance in avoiding human and dynamic obstacles, thereby ensuring safe navigation. The implementations are open source at: \url{https://github.com/thanhnguyencanh/SGan-TEB.git}
}


\maketitle


\section{Introduction}
\label{sec:introduction}
As the development of social capabilities in mobile service robots advances, effective navigation becomes crucial to autonomously performing tasks in dynamic unknown environments such as airports~\cite{shen2020sensor}, museums~\cite{hellou2022technical}, and offices~\cite{singh2023behavior}. The primary objective of these systems is to ensure that robot navigation is performed in a socially acceptable manner, prioritizing human safety and comfort~\cite{truong2019social}. However, in almost all scenarios, robots often face challenges in deciding how to proceed toward their intended destination unless an obstacle moves out of the way.

Several previous works have focused on designing obstacle avoidance algorithms and motion safety methods. These include artificial potential field~\cite{weerakoon2015artificial},  dynamic window approach (DWA)~\cite{fox1997dynamic}, vector field histogram~\cite{yim2014analysis}, velocity obstacles~\cite{wilkie2009generalized}, time elastic band (TEB)~\cite{rosmann2012trajectory}, randomized kinodynamic planning~\cite{bordalba2020randomized}, reciprocal velocity obstacles \cite{chen2024reciprocal}, and inevitable collision states~\cite{bautin2010inevitable}. Additionally, multi-sensors fusion methods~\cite{canh2022multisensor}, using camera and lidar sensors, have shown potential for effectively avoiding static and dynamic obstacles of various shapes and sizes in different environments. While these techniques have successfully avoided obstacles in dynamic environments, they often lack the human characteristics and social constraints necessary for friendly and safe navigation. 

To ensure the ability of navigation, Simultaneous Localization and Mapping~\cite{canh2023object}, \cite{canh2024s3m} approach is leveraged to localize and construct the surrounding environment. This approach provides the current robot pose and pre-estimated map, supporting navigation and obstacle avoidance tasks. However, human or dynamic obstacles cannot be defined on the static map, requiring a local planning algorithm capable of predicting their future motion during the robot's movement. On the other hand, computer vision and machine learning for robot navigation have inspired a significant amount of research over the past three decades~\cite{mavrogiannis2023core}. By investigating motion behavior prediction, mobile robots can effectively manage social interaction. Although existing human-aware robot navigation systems~\cite{truong2017socially},~\cite{nguyen2020proactive} based on the social force model have been developed and verified in the real world environment, achieving considerable success. However, these systems typically only consider human state information such as human position, orientation, and velocity.

In particular, interaction-aware planning has been advanced through the use of Partially Observable Markov Decision Processes~\cite{chen2023interaction} and Model Predictive Control~\cite{gupta2023interaction} with safety constraints. These methods provide robust frameworks for decision-making in uncertain environments by anticipating future states and optimizing control actions over a predicted horizon. Despite the strengths of these methods, there remains a gap in integrating these with real-time motion prediction and social adaptability in densely populated environments. 

To address these challenges, this paper proposes an effective approach by integrating motion prediction and trajectory planning for social robot navigation in dynamic human environments. We consider several conditions concerning social constraints and kinodynamic constraints. The main contributions of this work are summarized as follows:
\begin{itemize}
    \item An effective system for social robot navigation in dynamic human environments.
    \item Construction of a novel objective function for local planners based on social constraints.
    \item Investigation of human motion prediction based on Generative Adversarial Networks (GAN).
    \item Demonstration of our proposed system's performance in both quantitative and qualitative metrics.
\end{itemize}

The rest of this paper is organized as follows: Section 2 introduces our proposed system, focusing on human identification, motion estimation, and trajectory planning. Experimental results using both quantitative and qualitative metrics are shown in Section 3. Finally, Section 4 presents the conclusions of our system.


\begin{figure*}[!ht]
    \centering
    \includegraphics[width=0.95\linewidth]{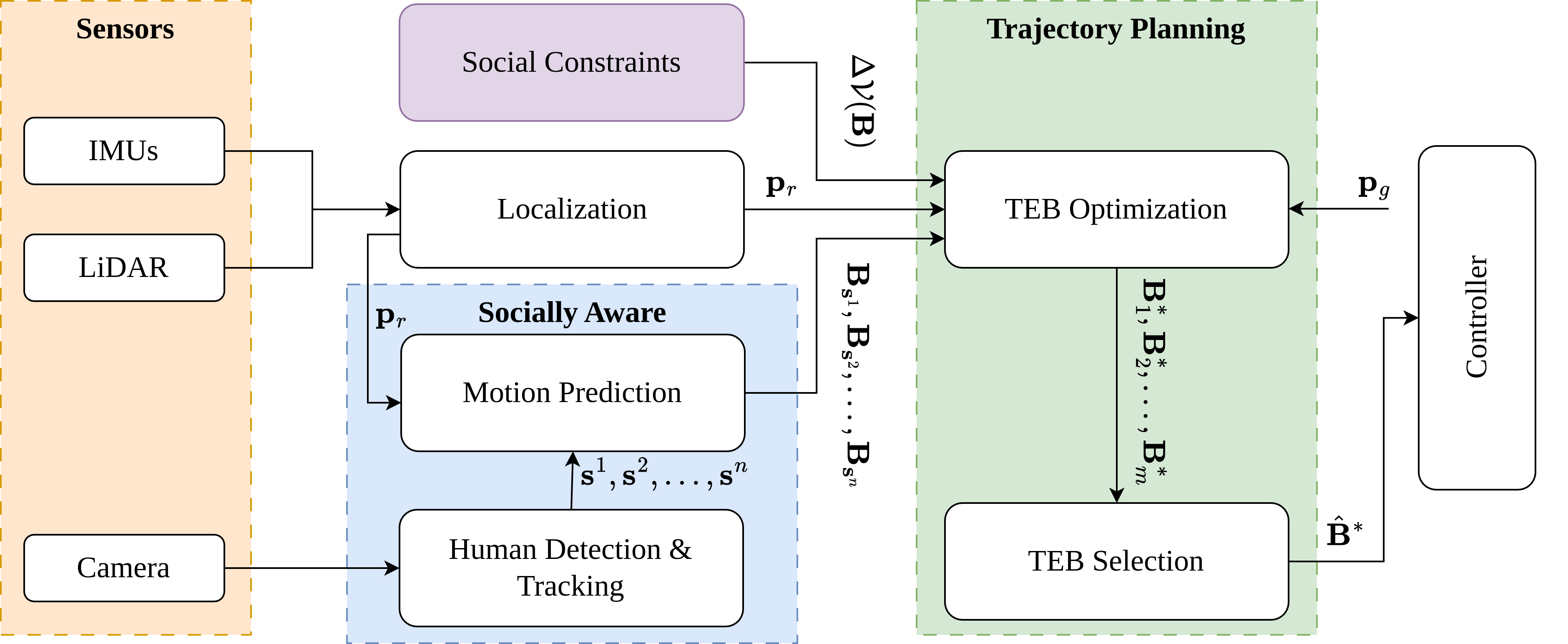}
    \caption{Overview of our proposed method. The system is composed of two main units: Socially Aware and Social Trajectory Planning}
    \label{fig:overview}
\end{figure*}

\section{Methodology}
\label{sec:proposed}
Our proposed pipeline is shown in Fig.~\ref{fig:overview}, which includes two main parts: socially aware and social trajectory planning. First, RGB-D images are used as input to detect and track humans, with the output being the human pose in world coordinates, while IMU and LiDAR data are used for localization path to estimate robot pose. After that, they are combined with robot poses to predict human trajectory by leveraging Social GAN method~\cite{gupta2018social}, which is detailed in Section~\ref{sec:social}\ ~. Subsequently, we construct a novel objective function by integrating social constraints such as personal distance and rules for moving within a narrow range into the TEB planner (Section~\ref{sec:socialNav} \ ~). 


\begin{figure}[!ht]
    \centering
    \includegraphics[width=0.8\linewidth]{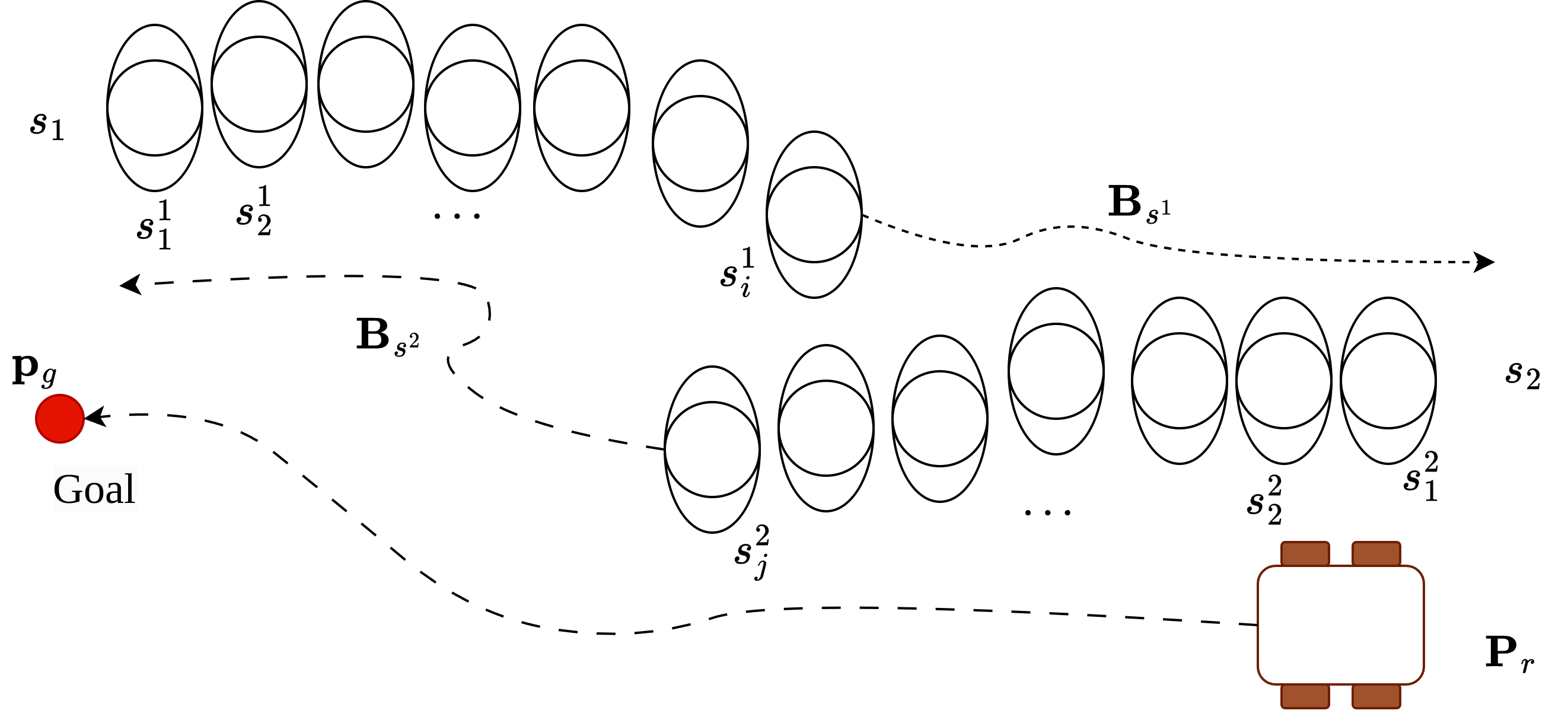}
    \caption{Illustrative example of social navigation in the dynamic human environment. The robot needs to achieve a predetermined goal while safely avoiding moving people $s_1$ and $s_2$.}
    \label{fig:problem}
\end{figure}

We consider robots as well as humans to aim to reach a pre-defined goal and safety navigation in a dynamic human environment like Fig.~\ref{fig:problem}. We denoted $\bm p_r = \begin{bmatrix}
    x_r & y_r & \theta_r
\end{bmatrix}^T$ as robot pose. 
$i-th$ person's trajectory is described by $\bm B_{\bm s^i} = (\bm s_1^i, \Delta T_1, \bm s_2^i, \Delta T_2, \dots, \bm s_n^i, \mathrm{\Delta T_N})$ with 
$\bm s^i_j = \begin{bmatrix}
    x_j^i & y_j^i
\end{bmatrix}^T$ is $i-th$ person position at the time $j$ and $\Delta T_k$ represents the time duration that the mobile robot
have to need to move between two consecutive poses $\bm s^i_k$ and $\bm s^i_{k+1}$. $\bm p_g = \begin{bmatrix}
    x_g & y_g & \theta_g
\end{bmatrix}^T$ is the pose of the destination. 

\subsection{Socially Aware} \label{sec:social}

The goal of the Socially Aware part is to estimate human pose and motion trajectory based on RGB-D images. We first detect humans and calculate human position following this equation:
\begin{equation}
    s^i = 
    \begin{bmatrix}
       x^i \\ y^i 
    \end{bmatrix} = \begin{bmatrix}
        \big (x^c_{min} + \frac{x^c_{max} - x^c_{min}}{2} - m_{cx} \big) \frac{d}{m_{fy}}\\
        d 
    \end{bmatrix}
\end{equation}
where $x^c_{min}, x^c_{max}$ are the minimum and maximum coordinate values along the $x-$axis of the human detection bounding box, respectively. $d$ is the depth value and $m_{cx}, m_{fy}$ are intrinsic camera parameters.

Afterward, we employed the Social GAN method~\cite{gupta2018social} to predict motion trajectory. The Social GAN consists of three main components: Generator (G), Pooling Module (PM), and Discriminator (D). G is based on the encoder-decoder conceptual framework, where the encoder and decoder hidden states are interconnected through PM. Given the input motion trajectory $\bar{\mathbf{B}_{\mathbf{s}^i}}$, G generates the predicted trajectory $\mathbf{B}_{\mathbf{s}^i}$. D evaluates the entire sequence including the input trajectory $\bar{\mathbf{B}_{\mathbf{s}^i}}$ and the generated prediction $\mathbf{B}_{\mathbf{s}^i}$, distinguishing between "real" and "fake" trajectories.

The choice of Social GAN is driven by its robust capability to model complex interactions and effectively predict nuanced human movement patterns, making it especially suitable for environments with unpredictable human behaviors. This method allows for a probabilistic rather than deterministic forecasting of future paths, enabling more flexible and adaptive navigation strategies. The trajectories of multiple individuals are tracked, capturing their $8$ previous positions recorded at intervals of 0$.5$ seconds. This data, along with human identifiers, is fed into the motion prediction system. Using Social GAN, future human trajectories are forecasted over the next $6$ seconds, comprising $12$ predicted positions along their trajectory. Subsequently, these predicted trajectories are integrated into the robot's local planning system for navigation and obstacle avoidance.

\subsection{Proposed MP-TEB algorithm} \label{sec:socialNav}

The main purpose of the MP-TEB method is to determine control commands that enable the robot to move from an initial pose to a goal pose within minimal time intervals while satisfying kinodynamic constraints, adhering to social constraints, and maintaining a safe distance from objects. To achieve this, we designed a new objective function that integrates social constraints and motion prediction information with the Time Elastic Band (TEB)~\cite{rosmann2012trajectory} method, a well-established motion planning approach. The traditional objective function $\mathcal{V}({\mathbf{B}})$ is defined as follows:

\begin{equation}
\begin{aligned}
    \mathcal{V}(\mathbf{B}) &= \sum_{\mathrm{i=1}}^{\mathrm{N-1}} \Big [\mathrm{\Delta T_i^2 + \delta_h} ||\mathbf{h}_{\mathrm{i}}||\mathrm{^2_2} + \delta_v ||\min \{\mathbf{0}, \mathbf{v}_{\mathrm{i}}\}||\mathrm{^2_2} \\
    &+ \delta_o ||\min\{\mathbf{0},\mathbf{o}_{\mathrm{i}}\}||\mathrm{^2_2} + \delta_{\alpha}||\min \{\mathbf{0}, \bm \alpha_{\mathrm{i}}\}||\mathrm{^2_2}  \Big ] \\
    &= \mathbf{w}^{\mathrm{T}} \mathbf{\textit{f}}(\mathbf{B})
\end{aligned}
\end{equation}
subject to:

$0 \leq \Delta T_i \leq \Delta T_{max}$

$\bm h_i(\bm s_{i+1}, \bm s_i) = 0$: Nonholonomic kinematics

$\bm o_i(\bm s_i )\geq 0$: Clearance from surrounding objects

$\bm v_i(\bm s_{i+1}, \bm s_i, \Delta T_i) \geq 0$: Limitation of robotic velocities

$\bm \alpha_i(\bm s_{i+1}, \bm s_i, \bm s_{i-1}, \Delta_i, \Delta T_{i-1}) \geq 0$: Robot accelerations limitation

The objective function in Eq. (2) aims to deform a path to the goal by applying internal constructive forces, which result in the shortest possible path, and external repulsive forces, which radiate from the obstacles to ensure a collision-free trajectory. In which $\bm w$ are individual weights corresponding to individual cost terms $f(\bm B)$. Constraints of the environment include kinematics, clearance from obstacles, and limitation of velocities and accelerations formula by equality and inequality equations.

The local optimal trajectory sequence with $m$ obstacles $(\bm B^*_1, \bm B^*_1, \dots, \bm B^*_m)$ is generated by solving the following non-linear program:

\begin{equation}
    \bm B^*_1, \bm B^*_1, \dots, \bm B^*_m = \argmin_{\bm B \setminus  \{\bm s_1, \bm s_N\}} \mathcal{V}(\mathbf{B})
\end{equation}

Then the optimal TEB trajectory $\hat{\bm B}^*$ based on the global minimum is determined by solving the following equations:
\begin{equation}
    \hat{\bm B}^* = \argmin_{\bm B^*_j \in \{\bm B^*_1, \bm B^*_1, \dots, \bm B^*_m \}}  \mathcal{V}(\mathbf{B}_{\mathrm{j}})
\end{equation}

In the proposed system, the robot's movement behavior is assumed as human. This means that the robot's navigation is like human navigation. This strategy fosters a natural interaction environment that allows humans to easily predict and respond to robot behavior and vice versa. To achieve this, the robot needs to create a path that closely follows the predicted path in human scenarios. It enhances predictability and smoothness in the robot’s movements, adapting paths that are intuitively safer for nearby humans. Therefore, a constraint that reflects a predicted human path can be formula based on the following cost function:
\begin{equation}
    e^{hum-like}_t = w_1 ||\bm p_{r_t} - \bm p_{s_t}||
\end{equation}
where $\bm p_{s_t}$ and $w_1$ are human pose and weight for human trajectory constraint at the time $t$, respectively.

The dynamic obstacle constraint cost is calculated as follows:

\begin{equation}
    e^{dyn-obs}_t = w_2 ||\bm p_{r_t} - (\bm p_{{obs}_t} + \bm v_{{obs}_t} t)||
\end{equation}
where $\bm p_{{obs}_t}$ and $\bm v_{{obs}_t}$ are dynamic obstacle pose and velocities at the time $t$, respectively. $w_2$ is dynamic obstacle constraint weight.

In addition, in narrow spaces such as corridors or entrances, a robot moving along the shortest path can cause obstruction or affect the movement space of surrounding people. Therefore, it is necessary for the robot to follow human movement rules. An important strategy is to move closer to the curb on the right-of-way and avoid people coming from the opposite direction. To determine whether the obstacle is on the priority side or the non-priority side of the robot, we use the following equation:

\begin{equation}
    e^{pri}_t = cos(\theta_r) (y_{{obs}_t} - y_{r_t}) - sin(\theta_{r_t}) (x_{{obs}_t} - x_{r_t})
\end{equation}
where $x_{{obs}_t}$ and $y_{{obs}_t}$ is position of obstacle at the time $t$. If $e^{pri}_t>0$, the obstacle is on the non-priority side of the robot and vice versa, the obstacle point is on the priority side of the robot.

Finally, we obtain the objective function of the proposed MP-TEB method as follows:
\begin{equation}
    \tilde{\mathcal{V}}(\bm B) = \mathcal{V}(\bm B) + \mathrm{\delta_{mp}(e^{hum-like}_t + e^{dyn-obs}_t - e^{pri}_t)}
\end{equation}
where, $\delta_{mp}$ is a normalization factor and are predefined value.

After achieving the optimal trajectory, the motion control command $\bm u_r = \begin{bmatrix}
    v_r, \omega_r
\end{bmatrix}^T$ is generated to control the mobile robot with $v_r$ and $\omega_r$ is linear velocity and angular velocity, respectively. 
The kinematic model is given by:

\begin{equation}
    \begin{bmatrix}
        x_r^{i+1} \\
        y_r^{i+1} \\
        \theta_r^{i+1} 
    \end{bmatrix} = \begin{bmatrix}
        x_r^{i} \\
        y_r^{i} \\
        \theta_r^{i} 
    \end{bmatrix} + \begin{bmatrix}
        \frac{v_r^r + v_r^l}{2} \cos{\theta_i}d_t \\
        \frac{v_r^r + v_r^l}{2} \sin{\theta_i}d_t \\
        \frac{v_r^r - v_r^l}{L}d_t 
    \end{bmatrix}
\end{equation}
where $v_r^r = v_r + \frac{L \omega_r}{2}d_t$ and $v_r^l = v_r - \frac{L \omega_r}{2}d_t$ are the velocities of the robot's right and left wheels, respectively, $L$ is the wheelbase of the robot.
 

\begin{figure}[!ht]
    \centering
    \begin{subfigure}[b]{0.125\textwidth}
    \centering
    \includegraphics[width=\textwidth]{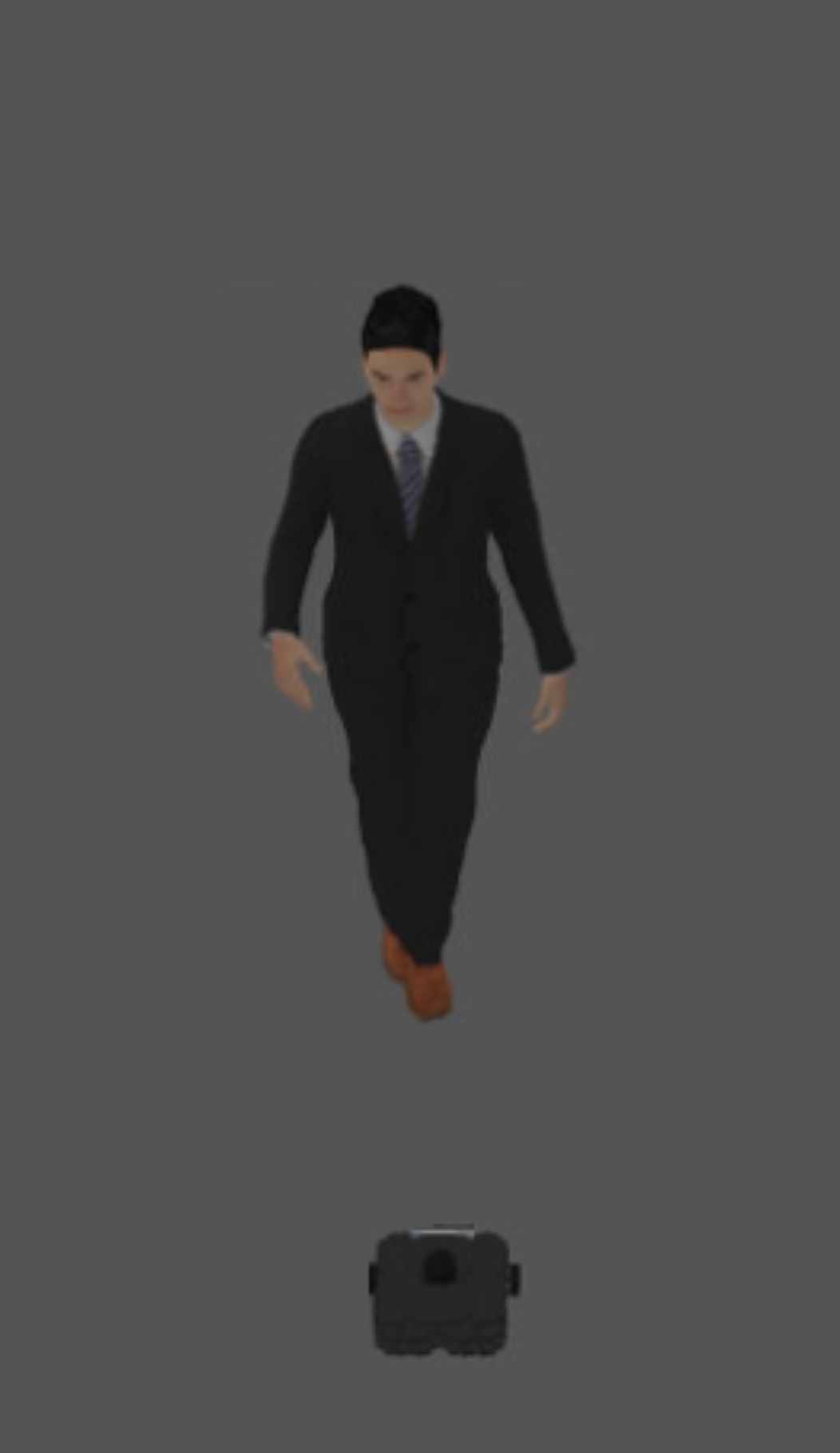}
    \caption{}
    \end{subfigure}
    \begin{subfigure}[b]{0.175\textwidth}
    \centering
    \centering
    \includegraphics[width=\textwidth]{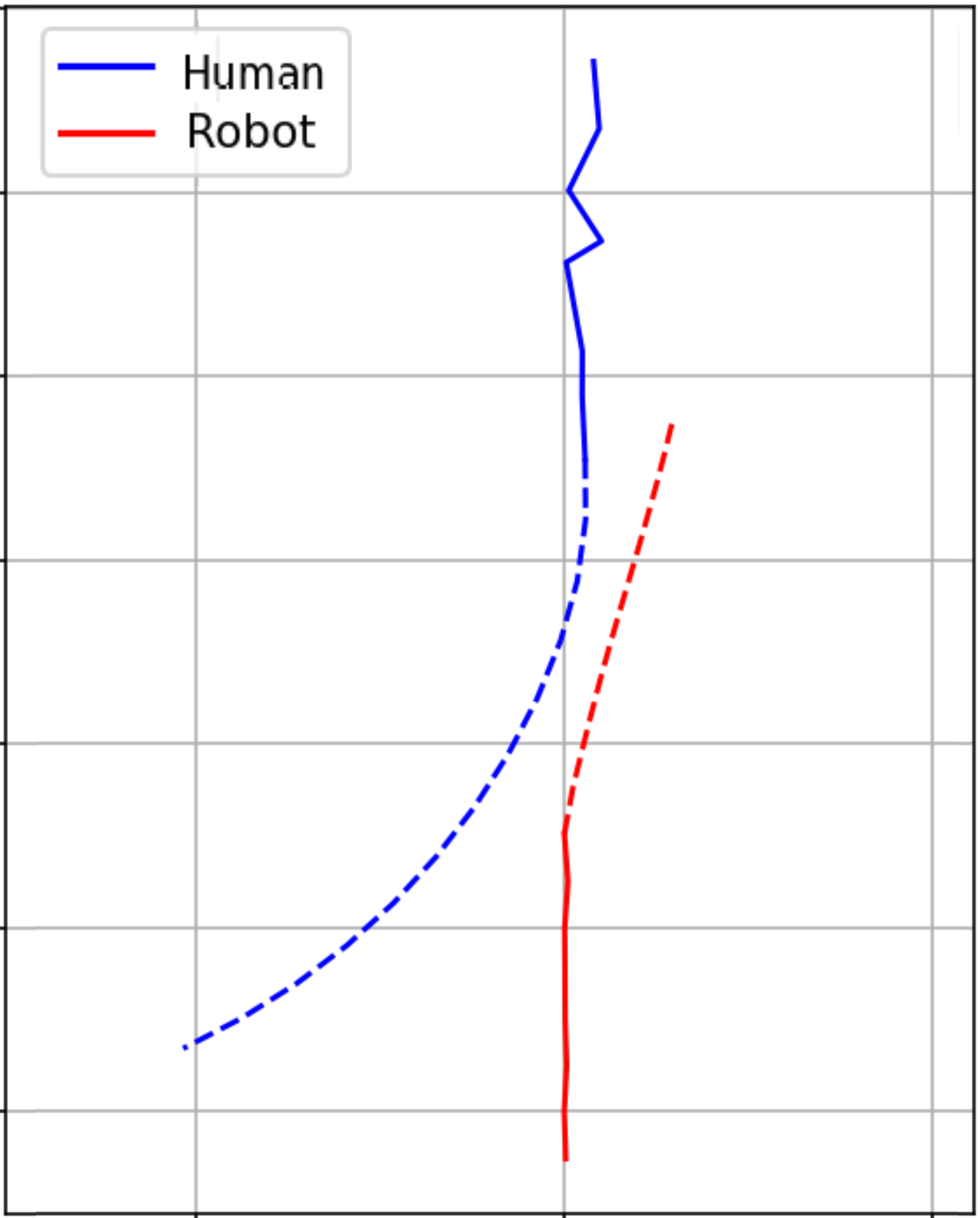}
    \caption{}
    \end{subfigure}
    \begin{subfigure}[b]{0.15\textwidth}
    \centering
    \centering
    \includegraphics[width=\textwidth]{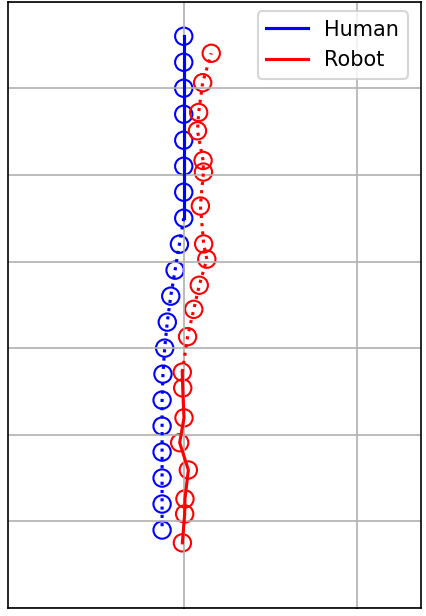}
    \caption{}
    \end{subfigure}
    \caption{Scenario 1: Reverse direction (a) simulation environment, (b) trajectory estimation, (c) real trajectory.}
    \label{fig:scen1}
\end{figure}

\begin{figure*}[!ht]
    \centering
    \begin{subfigure}[b]{0.28\textwidth}
    \centering
    \includegraphics[width=\textwidth]{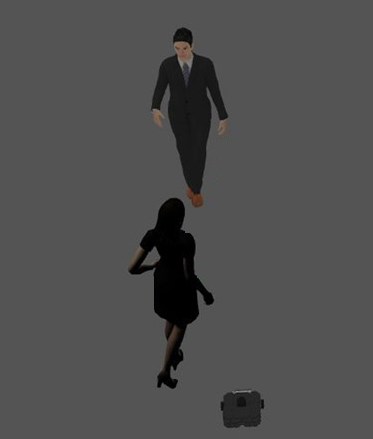}
    \caption{}
    \end{subfigure}
    \begin{subfigure}[b]{0.229\textwidth}
    \centering
    \centering
    \includegraphics[width=\textwidth]{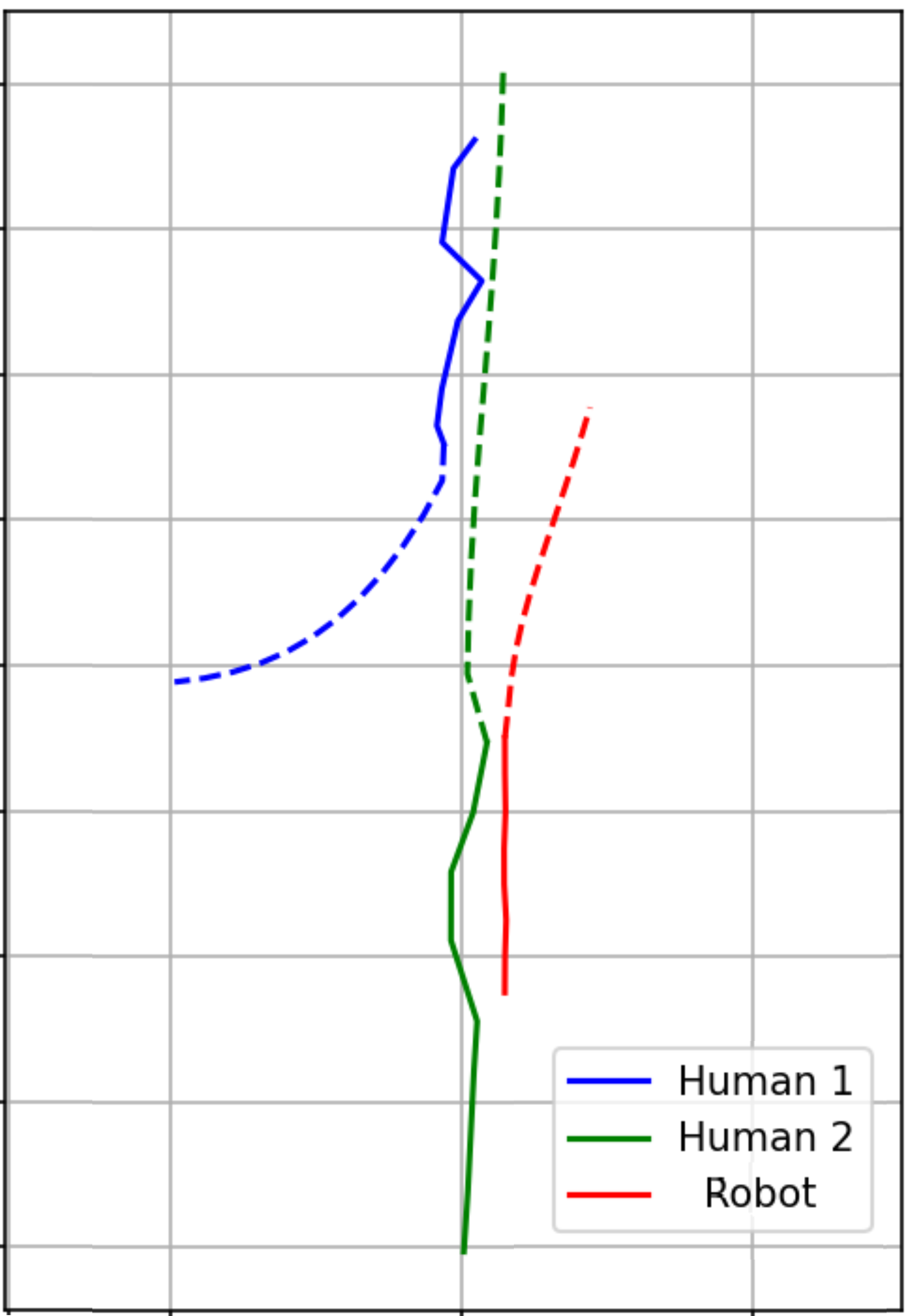}
    \caption{}
    \end{subfigure}
    \begin{subfigure}[b]{0.304\textwidth}
    \centering
    \centering
    \includegraphics[width=\textwidth]{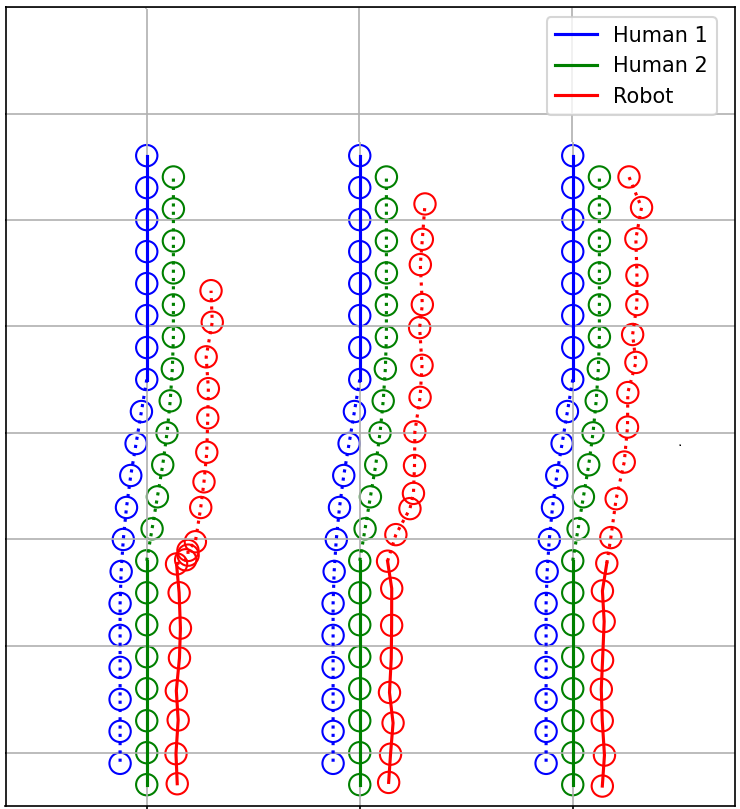}
    \caption{}
    \end{subfigure}
    \caption{Scenario 2: Multi-person (a) simulation environment, (b) trajectory estimation, (c) real trajectory (left: DWA, mid: TEB, right: MP-TEB).}
    \label{fig:scen2}
\end{figure*}

\section{Experiments}
\label{sec:result}

To demonstrate the capabilities of our proposed system, we conducted extensive tests within a physical simulation environment. We utilized two frameworks, the Gazebo simulator and the Robot Operating System (ROS), to create the environment and visualize the results. The software for the proposed system was developed using ROS-C++ and ROS-Python.

\subsection{Simulaion Setup}
The robot is set to move from the start position to the pre-defined goal position. The robot's initial velocity is set to $0.0 [m/s]$ and the maximum linear and angular velocities set to $1.0[m/s]$ and $0.5[rad/s]$, respectively. The range of the LiDAR sensor is $8 [m]$ and the camera has a horizontal field of view (FOV) of $87$ degrees.

\begin{table*}[htbp]
\centering
\caption{Quantitative results of social robot navigation.}
\begin{tabular}{p{0.15\textwidth}||p{0.06\textwidth}|p{0.06\textwidth}|p{0.07\textwidth}||p{0.06\textwidth}|p{0.06\textwidth}|p{0.07\textwidth}||p{0.06\textwidth}|p{0.06\textwidth}|p{0.07\textwidth}}
\toprule[0.5pt]
\multirow{2}{*}{Scenarios} & \multicolumn{3}{c}{DWA~\cite{fox1997dynamic}} & \multicolumn{3}{c}{TEB~\cite{rosmann2012trajectory}} & \multicolumn{3}{c}{MP-TEB} \\ 
& \textit{Path Length (m)} & \textit{Total Time (s)} & \textit{Min H-R Dist (m)} & \textit{Path Length (m)} & \textit{Total Time (s)} & \textit{Min H-R Dist (m)} & \textit{Path Length (m)} & \textit{Total Time (s)} & \textit{Min H-R Dist (m)}  \\
\midrule[0.5pt]
Reverse direction & 8.01 & 24.04 & 0.22 & 7.89 & 22.76 & 0.47 & \textbf{7.23} & \textbf{20.01} & \textbf{0.64} \\
Multi persons     & 8.11 & 26.33 & 0.29 & 7.57 & 25.16 & \textbf{0.46} & \textbf{6.73} & \textbf{23.80} & 0.36 \\
Corridor          & 7.75 & \textbf{23.12} & 0.40 & \textbf{6.95} & 24.94 & 0.39 & 7.11 & 24.46 & \textbf{0.79} \\
Turn right        & 6.97 & 23.12 & 0.34 & 7.07 & 22.62 & 0.29 & \textbf{6.54} & \textbf{21.38} & \textbf{0.83} \\
\bottomrule[0.5pt]
\end{tabular}
\label{tab:1}
\end{table*}

To evaluate the performance of our MT-TEB algorithm, we created four typical scenarios including (1) Scenario 1 - a mobile robot and a person moving in the reverse direction, (2) Scenario 2 - a mobile robot avoiding two persons, (3) Scenario 3 - a mobile robot move around the door in the narrow corridor, (4) Scenario 4 - a robot turn right and avoid a person. For each scenario, we conducted three experiments corresponding to a different trajectory planning algorithm: DWA~\cite{fox1997dynamic}, TEB~\cite{rosmann2012trajectory}, and MP-TEB. Experimental results are compared based on qualitative metric and quantitative metrics, which include 
\begin{itemize}
    \item Path Length $l$ (m) is the total length of the trajectory the robot has moved
    \item Total time $t$ (s) is the total time consumed when the robot moves from start to goal
    \item Min H-R distance $d$ (m) is the closest distance in the moving trajectory between the human and the robot
\end{itemize}
\subsection{Results}

\begin{figure}[ht]
    \centering
    \begin{subfigure}[b]{0.2\textwidth}
    \centering
    \includegraphics[width=\textwidth]{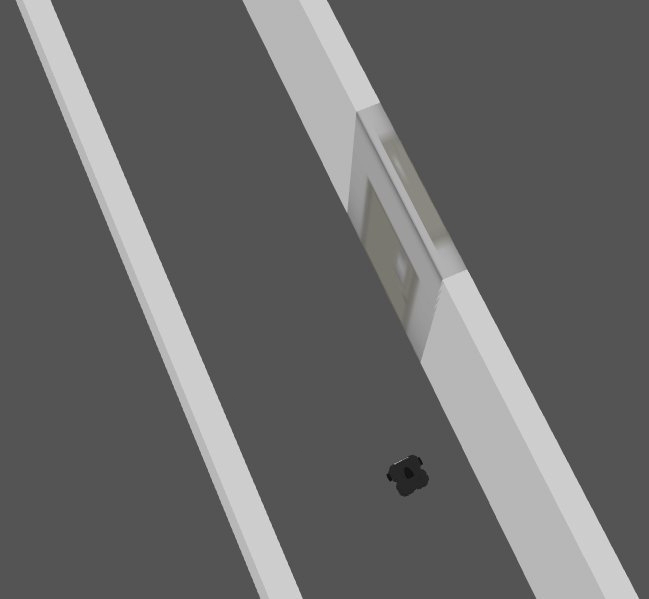}
    \caption{}
    \end{subfigure}
    \begin{subfigure}[b]{0.2\textwidth}
    \centering
    \centering
    \includegraphics[width=\textwidth]{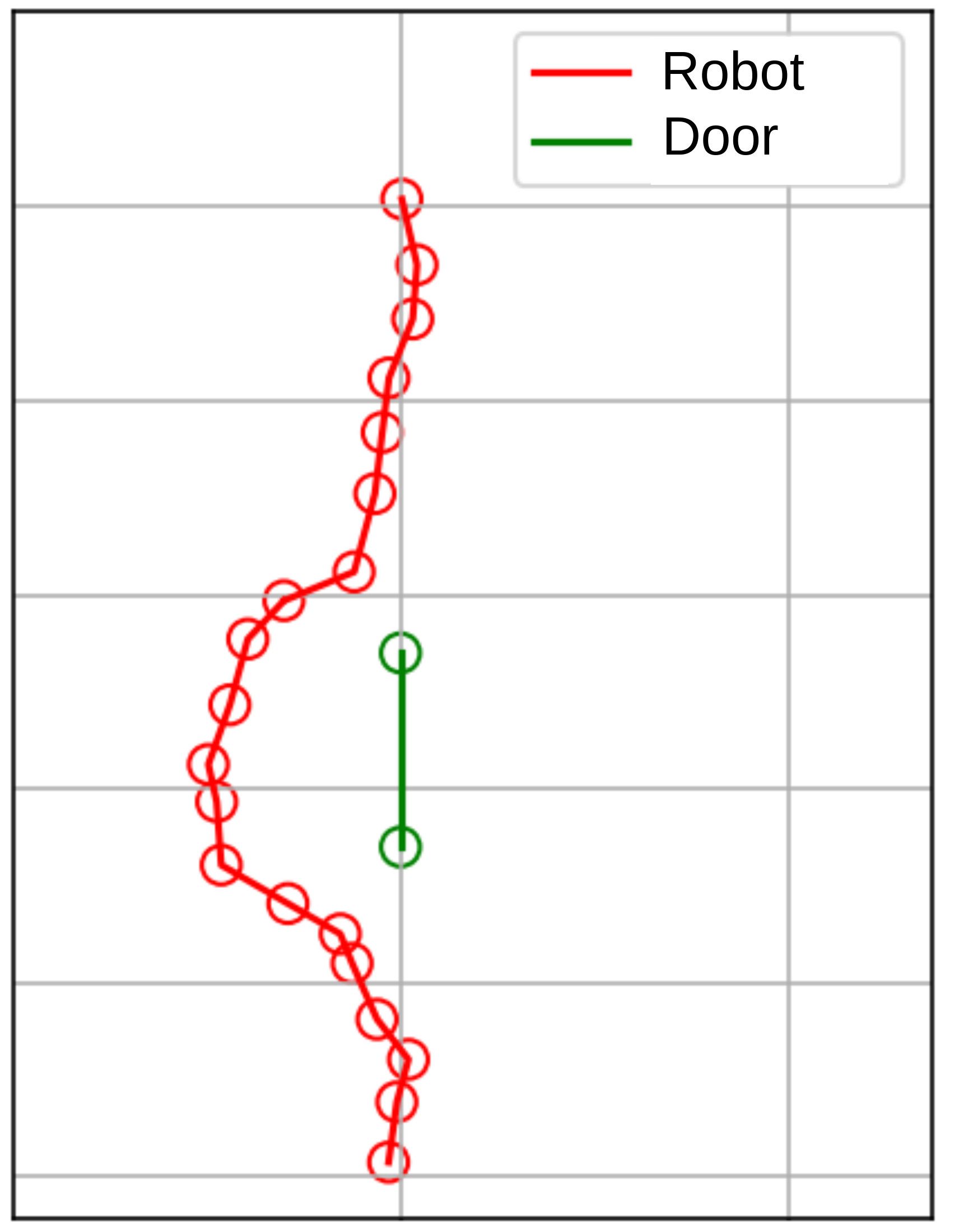}
    \caption{}
    \end{subfigure}
    \caption{Scenario 3: Avoid door in corridor scenario: (a) simulation environment, (b) real trajectory.}
    \label{fig:scen3}
\end{figure}

\begin{figure*}[ht]
    \centering
    \begin{subfigure}[b]{0.28\textwidth}
    \centering
    \includegraphics[width=\textwidth]{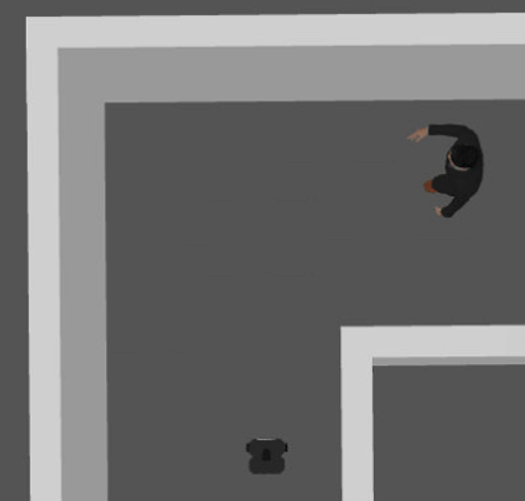}
    \caption{}
    \end{subfigure}
    \begin{subfigure}[b]{0.292\textwidth}
    \centering
    \centering
    \includegraphics[width=\textwidth]{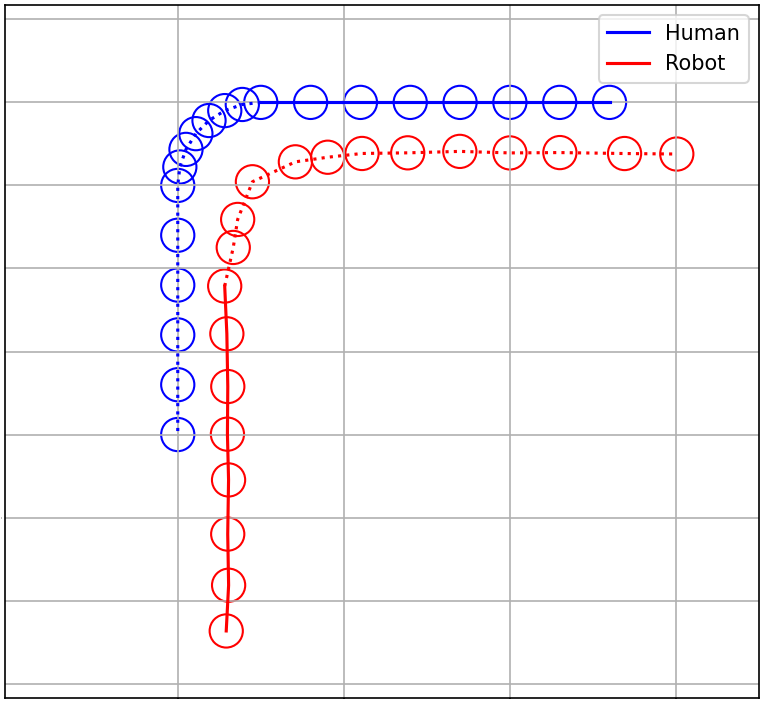}
    \caption{}
    \end{subfigure}
    \begin{subfigure}[b]{0.288\textwidth}
    \centering
    \centering
    \includegraphics[width=\textwidth]{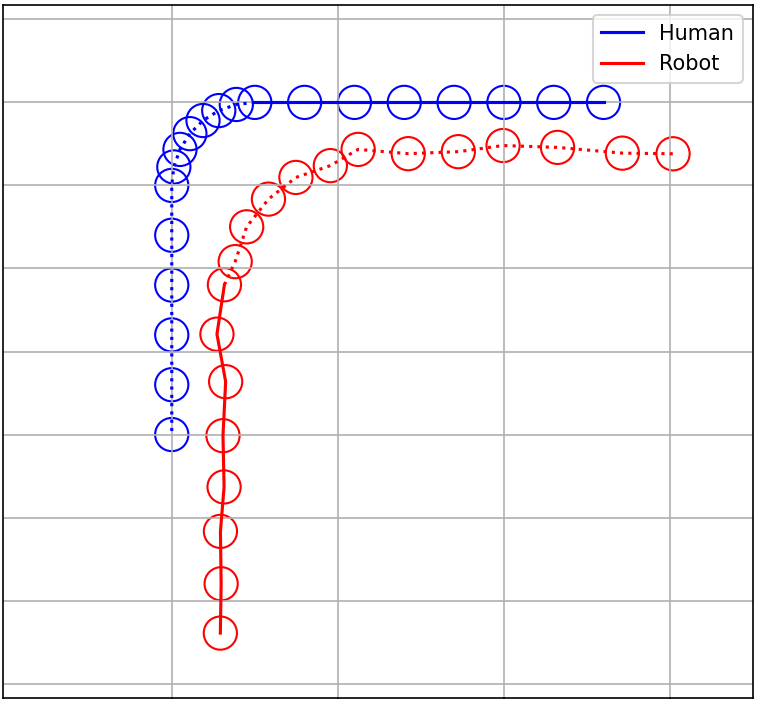}
    \caption{}
    \end{subfigure}
    \caption{Scenario 4: Turn right (a) simulation environment, (b) TEB algorithm, (c) MP-TEB algorithm.}
    \label{fig:scen4}
\end{figure*}

For the robot's movements to be human-compatible, the trajectory predictor processes both the robot's trajectory and the human's trajectory to create a dynamic human-robot environment. The actual trajectory is illustrated with a solid line, while the predicted trajectory is represented by a dashed line, as shown in Fig.~\ref{fig:scen1}b, Fig.~\ref{fig:scen2}b. Fig.~\ref{fig:scen1}, Fig.~\ref{fig:scen2} illustrate the results of Scenario 1 and Scenario 2, respectively. Using data from motion prediction, the robot successfully navigates around the predicted movement area of two people in the test environment. This proactive response enhances safety by allowing the robot to anticipate and avoid potential surprises. A qualitative comparison of DWA, TEB, and MP-TEB is presented in Fig~\ref{fig:scen2}c. It can be seen that with the proposed system, the path created is smoother and more natural due to its superior ability to predict future human positions. In contrast, the DWA method is surprised by the human's reaction to avoid the road, causing its movement to stop a lot, because this method fails to anticipate the movement of dynamic obstacles. Although the TEB method can avoid obstacles, it is typically a less smooth and natural path compared to our proposed method. 

In Scenario 3, the robot was able to safely move away from the closed door, as illustrated in Fig.~\ref{fig:scen3}. In Scenario 4, the robot turns a wide angle at the intersection for the conventional TEB method, which risks causing a sudden collision with a human because it does not maintain a safe distance enough as shown in Fig.~\ref{fig:scen4}b. For the proposed MP-TEB method, the robot turns a smaller angle, creating greater safety, while saving robot distance and travel time (Fig.~\ref{fig:scen4}c)

To evaluate the stable performance of the MP-TEB method, we repeated it 10 times in each scenario and averaged the results as shown in Table~\ref{tab:1}. In scenarios such as navigating in the reverse direction, DWA achieved a path length of $8.01 m$, a total time of $24.04 s$, and a minimum distance of $0.22 m$, while MP-TEB improved to $7.23 m$, $20.01 s$, and $0.64 m$, respectively. The maneuvering between multiple people showed that MP-TEB had the shortest path length ($6.73 m$), the lowest total time ($23.80 s$), and the safe minimum distance ($0.36 m$) compared to DWA and TEB. Navigating through narrow corridors, MP-TEB maintained competitive path length ($7.11 m$), total time ($24.46 s$), and safe distance ($0.79 m$). When performing a right turn to avoid a human, MP-TEB achieved superior performance with a path length of $6.54 m$, a total time of $21.38 s$, and a safe minimum distance of $0.83 m$, outperforming DWA and TEB in both efficiency and safety metrics in all scenarios. Overall, the total time of the MP-TEB method is much shorter than those of the other two methods because the path of the MP-TEB method is optimized for humans, so it does not take much time for the system to adapt to changes in human movement.


\section{Conclusion}
\label{sec:conclusion}

In conclusion, our study introduces and evaluates the Motion-Predictive Timed Elastic Band (MP-TEB) algorithm to enhance the navigation capabilities of mobile service robots in dynamic human environments. Through simulations and quantitative assessments across varied scenarios, including complex maneuvers and obstacle avoidance tasks, MP-TEB consistently outperformed traditional methods like the Dynamic Window Approach (DWA) and the standard Timed Elastic Band (TEB). The algorithm demonstrated superior performance in terms of path efficiency, navigation time, and maintaining safe distances from humans, ensuring effective and socially acceptable robot interactions. However, limitations include the need for further refinement in real-world deployment scenarios and adapting to highly unpredictable human behaviors. Moving forward, further refinements and integration of advanced technologies could continue to enhance MP-TEB's capabilities, contributing to its broader adoption in real-world applications where robots interact closely with humans.

\section*{Acknowledgments}
The authors would like to thank Quan Tien Bui, a former bachelor student at VNU-UET for his support in the development of the simulation environment and software during the study of this work.


\bibliographystyle{IEEEtran}
\bibliography{ref}  

\end{document}